\documentclass[10pt,twocolumn,letterpaper]{article}

\usepackage{cvpr}
\usepackage{times}
\usepackage{epsfig}
\usepackage{graphicx}
\usepackage{amsmath}
\usepackage{amssymb}
\usepackage{caption}
\usepackage{multirow}
\usepackage{subfig}
\usepackage{comment}
\usepackage{booktabs}
\usepackage[ruled,vlined]{algorithm2e}
\usepackage{microtype}      
\usepackage{alphalph}

\usepackage[pagebackref=true,breaklinks=true,letterpaper=true,colorlinks,bookmarks=false]{hyperref}
\usepackage{CJKutf8}

\usepackage[moderate,mathdisplays=normal,wordspacing=normal]{savetrees}

\cvprfinalcopy 
\newcommand{\tnote}[3]{{\color{#2}#1: #3}}

\newcommand{\seung}[1]{\tnote{SEUNG}{blue}{#1}}
\newcommand{\IGNORE}[1]{}

\usepackage{mathtools}
\usepackage{xspace}

\newcommand{\mbc}{\mathbf{c}}

\newcommand{\mbh}{\mathbf{h}}
\newcommand{\mbj}{\mathbf{j}}
\newcommand{\mbm}{\mathbf{m}}

\newcommand{\mbp}{\mathbf{p}}

\newcommand{\mbs}{\mathbf{s}}

\newcommand{\mbv}{\mathbf{v}}

\newcommand{\mbx}{\mathbf{x}}
\newcommand{\mby}{\mathbf{y}}
\newcommand{\mbz}{\mathbf{z}}

\newcommand{\R}{\mathbb{R}}

\newcommand{\calJ}{\mathcal{J}}

\newcommand{\calM}{\mathcal{M}}

\newcommand{\calV}{\mathcal{V}}

\newcommand{\calX}{\mathcal{X}}
\newcommand{\calY}{\mathcal{Y}}
\newcommand{\calZ}{\mathcal{Z}}

\def\R{\mathbb{R}}

\def\calX{\mathcal{X}}

\newcommand{\ignore}[1]{}

\makeatletter
\DeclareRobustCommand\onedot{\futurelet\@let@token\@onedot}
\def\@onedot{\ifx\@let@token.\else.\null\fi\xspace}
\def\eg{{e.g}\onedot} 
\def\ie{{i.e}\onedot} 

\def\etal{{et al}\onedot}
\makeatother

\def\cvprPaperID{55} 

\ifcvprfinal\fi
\begin{document}
\title{Pushing the Envelope for RGB-based \\
Dense 3D Hand Pose Estimation via Neural Rendering}

\author{Seungryul Baek\\
Imperial College London\\
{\tt\small s.baek15@imperial.ac.uk}
\and
Kwang In Kim\\
UNIST\\
{\tt\small kimki@unist.ac.kr}
\and
Tae-Kyun Kim\\
Imperial College London\\
{\tt\small tk.kim@imperial.ac.uk}
}

\maketitle
\thispagestyle{empty}

\begin{abstract}
Estimating 3D hand meshes from single RGB images is challenging, due to intrinsic 2D-3D mapping ambiguities and limited training data. We adopt a compact parametric 3D hand model that represents deformable and articulated hand meshes. To achieve the model fitting to RGB images, we investigate and contribute in three ways: 1) Neural rendering: inspired by recent work on human body, our hand mesh estimator (HME) is implemented by a neural network and a differentiable renderer, supervised by 2D segmentation masks and 3D skeletons. HME demonstrates good performance for estimating diverse hand shapes and improves pose estimation accuracies. 2) Iterative testing refinement: Our fitting function is differentiable. We iteratively refine the initial estimate using the gradients, in the spirit of iterative model fitting methods like ICP. The idea is supported by the latest research on human body. 3) Self-data augmentation: collecting sized RGB-mesh (or segmentation mask)-skeleton triplets for training is a big hurdle. Once the model is successfully fitted to input RGB images, its meshes i.e. shapes and articulations, are realistic, and we augment view-points on top of estimated dense hand poses. Experiments using three RGB-based benchmarks show that our framework offers beyond state-of-the-art accuracy in 3D pose estimation, as well as recovers dense 3D hand shapes. Each technical component above meaningfully improves the accuracy in the ablation study. 
\end{abstract}

\section{Introduction}
\begin{figure}[!t]
\captionsetup[subfigure]{labelformat=empty}
\centering
\subfloat[]{\includegraphics[width=0.23\linewidth]{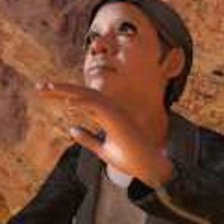}}
\subfloat[]{\includegraphics[width=0.23\linewidth]{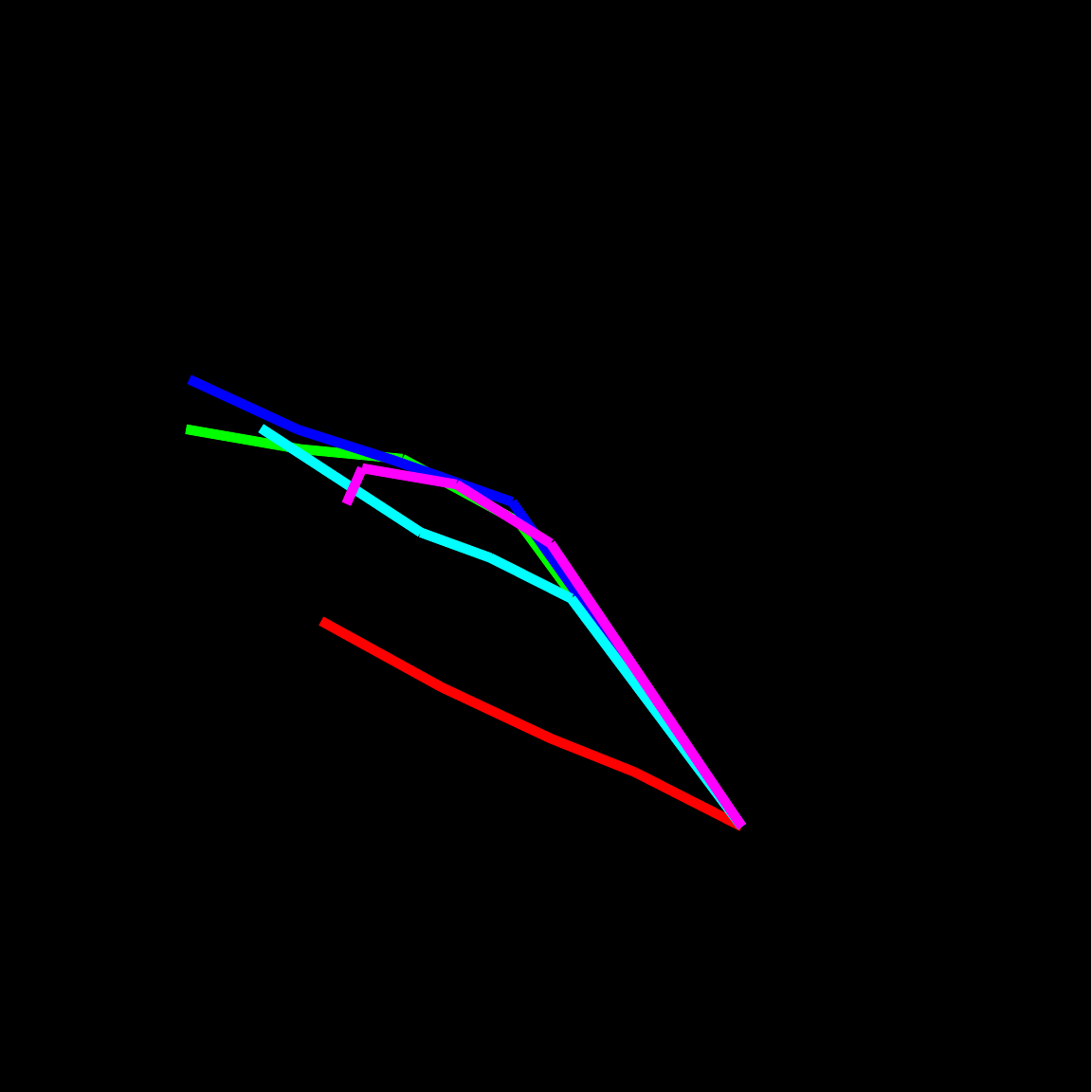}}
\subfloat[]{\includegraphics[width=0.23\linewidth]{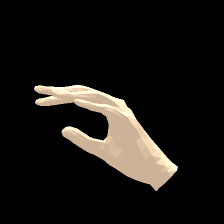}}
\subfloat[]{\includegraphics[width=0.23\linewidth]{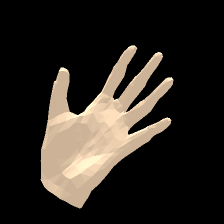}}\\
\vspace{-0.86cm}
\subfloat[]{\includegraphics[width=0.23\linewidth]{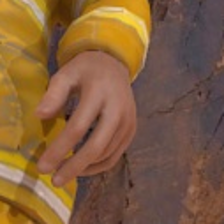}}
\subfloat[]{\includegraphics[width=0.23\linewidth]{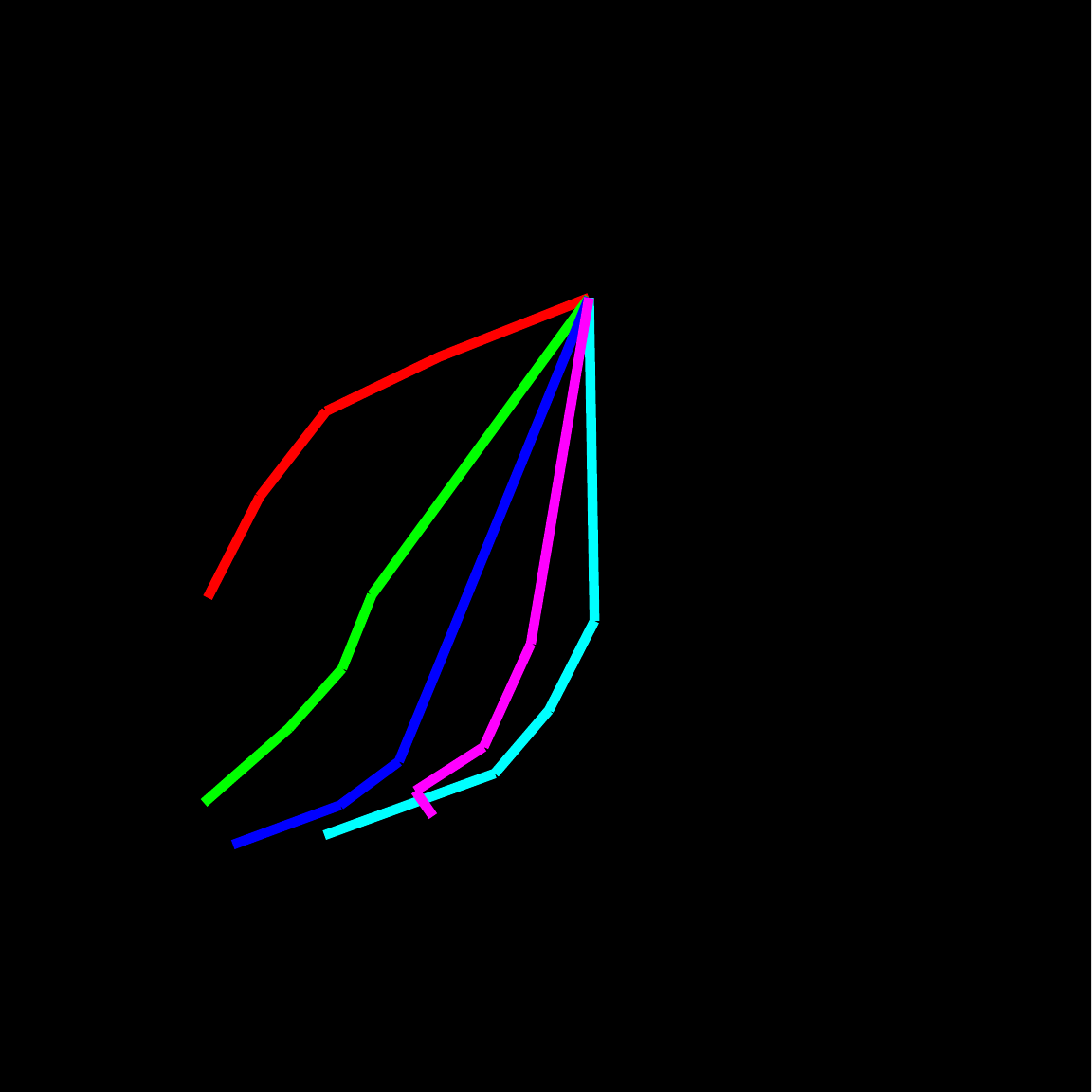}}
\subfloat[]{\includegraphics[width=0.23\linewidth]{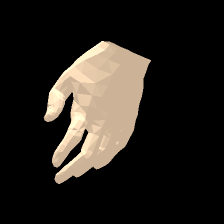}}
\subfloat[]{\includegraphics[width=0.23\linewidth]{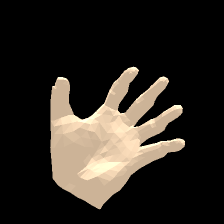}}
\vspace{-5mm}
\caption{Dense hand pose estimation examples. Our system estimates 3D shapes, as well as articulations and viewpoints. Left to right: input images, \emph{coarse} skeletal representations, \emph{dense} hand pose representations, and recovered hand shapes in a canonical articulation/viewpoint. Dense pose estimation provides a richer description of hands and improves the pose estimation accuracy.}
\label{fig:first}
\end{figure}

Recovering hand poses and shapes from images enables many real-world applications, \eg hand gesture as a primary interface for AR/VR. The problem is challenging due to high dimensionality of hand space, pose and shape variations, self-occlusions, etc~\cite{hrf_iccv2013,hand_cvpr_2017_1,hand_cvpr_2017_2,hand_cvpr_2017_3,hand_cvpr_2017_4,hand_bmvc_2017_1,hand_bmvc_2017_2,hand_icra_2017,hand_icip_2017,hand_eccv_2016_1,hand_eccv_2016_2,hand_eccv_2016_3,hand_icpr_2016,hand_ijcai_2016,hand_siggraphasia_2016,hand_siggraph_2016,hand_cvpr_2016_1,hand_cvpr_2016_2,hand_cvpr_2016_3,hand_cvpr_2016_4,deepprior,keskin2012,melaxi3d,forth,gui_cvpr_2018,handpose2_cvpr2018}. Most existing methods have focused on recovering sparse hand poses i.e. skeletal articulations from either a depth or RGB image. However, estimating dense hand poses including 3D \emph{shapes} (Fig.~\ref{fig:first}) is important as it helps understand e.g. human-object interactions~\cite{gui_cvpr_2018,iccv17_objecthand,seungryul2,forestwacv} and perform robotic grasping, where surface contacts are essential.

Discriminative methods based on convolutional neural networks (CNNs) have shown very promising performance in estimating 3D hand poses either from RGB images~\cite{hand_cvpr_2017_1,iccv_2017_zimmerman,eccv2018_rgbhand,eccv2018_rgbhand2,ganerated_cvpr18,cvpr_2018_crossmodal} or depth maps~\cite{hand_cvpr_2017_3,hand_iccvw_2017_1,hand1,hand_cvpr_2017_2,hand_iccvw_2017_1,handpose2_cvpr2018,hand_cvpr18_v2v,hand_cvpr18_markus,handpose2_cvpr2018,baek_cvpr2018}.
However, the predictions are based on coarse skeletal representations, and no explicit kinematics and geometric mesh constraints are often considered. On the other hand, establishing a personalized hand model requires a generative approach that optimizes the hand model to fit to 2D images~\cite{hand_bmvc_2017_2,forth,cvpr14_qian,hand_iccv_2015_1,hand_cvpr_2016_4,handmodel_cvpr_2014}. Optimization-based methods, besides their complexity, are susceptible to local minima and the personalized hand model calibration contradicts the generalization ability for hand shape variations.

Sinha~\etal~\cite{cvpr2017_surfnet} learn to generate 3D hand meshes from depth images. A direct mapping between a depth map and its 3D surface is learned; however, the accuracy shown is limited. Malik~\etal~\cite{3dv_2018_deephps} incorporate a 3D hand mesh model to CNNs and learn a mapping between depth maps and mesh model parameters. They further raycast 3D meshes with various viewpoint, shape and articulation parameters to generate a million-scale dataset for training CNNs. Our work tackles the similar but in RGB domain and offers an iterative mesh fitting using a differentiable renderer~\cite{cvpr_2018_neuralrenderer}. This improves once estimated mesh parameters using the gradient information at testing. Our framework also allows indirect 2D supervisions (2D segmentation masks, keypoints) instead of using 3D mesh data. 

While 2D keypoint detection is well established in the RGB domain~\cite{hand_cvpr_2017_1}, estimating 3D keypoints from RGB images is less trivial. Recently, several methods were developed~\cite{iccv_2017_zimmerman,wacv2018_hand,eccv2018_rgbhand,eccv2018_rgbhand2}. While directly lifting 2D estimations to 3D was attempted in~\cite{iccv_2017_zimmerman}, 2.5D depth maps are estimated as clues for 3D lifting in state-of-the-art techniques~\cite{eccv2018_rgbhand,eccv2018_rgbhand2}. In this paper, we exploit a deformable 3D hand mesh model, which inherently offers a full description of both hand shapes and articulations, 3D priors for recovering depths, and self-data augmentation. Different from purely generative optimization methods \eg~\cite{wacv2018_hand}, we propose a method based on a neural renderer and CNNs. 

Our contribution is in implementing a full 3D dense hand pose estimator (DHPE) using single RGB images via a neural renderer. Our technical contributions are largely threefold: 

1) DHPE is composed of convolutional layers that first estimate 2D evidences (RGB features, 2D keypoints) from an RGB image and then estimate the 3D mesh model parameters. Since RGB/mesh pairs are lacking, the network is learned to fit the 3D model using 2D segmentation masks and skeletons as supervision, similar to recent work on human body pose estimation~\cite{nips_2017_motioncapture,e2eshapepose,bodyrecon_cvpr18}, via neural renderer. The dense shape estimation helps improve the pose estimation. 

2) At the testing time, we iteratively refine the initial 3D mesh estimation using the gradients. The gradients are computed over self-supervisions by comparing estimated 3D meshes to predicted 2D evidences, as ground-truth labels are not available at testing. While previous work~\cite{nips_2017_motioncapture} fixed 2D skeletons/segmentation masks during the refinement step, we recursively improve 2D skeletons and exploit the improved skeletons and feature similarities for 2D masks. 

3) To further deal with limited annotated training data, especially for diverse shapes and view-points, we supply fitted meshes and their 2D projected RGB maps by varying shapes and view-points, to fine-tuning the network. Purely synthetic data imposes a synthetic-real domain gap, and annotating 3D meshes or 2D segmentation masks of real images is difficult. Once the model is successfully fitted to input RGB images, its meshes, \ie shapes and poses are close-to-real.

The proposed method can also be seen as a hybrid method~\cite{FG_hybrid_2015,hand_eccv_2016_1,hand_hybrid_2} combining merits of discriminative and generative approaches. 

\section{Related work}

\noindent\textbf{Hand pose estimation.} 
There have been several categories of works predicting hand poses. They vary according to types of output (3D mesh, 3D skeletal and 2D skeletal representations) and input (RGB and depth maps). 

\emph{3D hand model fitting methods.} In~\cite{hand_siggraphasia_2016,hand_siggraph_2016,mano,hand_iccv_2015_1,hand_cvpr_2016_4,handmodel_cvpr_2014,hand_bmvc_2017_2,forth}, 3D hand mesh representation is used to describe hand poses. To recover 3D meshes from 2D images, most methods rely on complex optimization methods or use point clouds to fit their 3D mesh models. One recent method~\cite{wacv2018_hand} takes RGB images as input and formulates the framework that runs in real-time. Their 3D mesh model is matched to estimated 2D skeletons. However, 2D skeleton is coarse and overall 3D fitting fails when the 2D skeleton estimation is not accurate.

\emph{Recovering 3D skeletal representations from depth images.} 
Estimating 3D hand skeletons from depth images has shown promising accuracies~\cite{handpose2_cvpr2018}. The problem is better set since the depth input inherently offers rich 3D information~\cite{hrf_iccv2013}. Further, deep learning methods~\cite{hand_cvpr_2017_3,hand_iccvw_2017_1,hand1,hand_cvpr_2017_2,hand_iccvw_2017_1,handpose2_cvpr2018,hand_cvpr18_v2v,hand_cvpr18_markus,handpose2_cvpr2018,baek_cvpr2018} have been successfully applied along with the million-scale hand pose dataset \cite{hand_cvpr_2017_3} in this domain. Recently, Malik~\etal~\cite{3dv_2018_deephps} proposed to estimate both 3D skeletal and mesh representations and have obtained improved accuracy in 3D skeletal estimation. They collect a new million-scale data set for depth map and 3D mesh pairs to train the network in a fully supervised manner. 

\begin{figure*}[t]
\includegraphics[width=1\linewidth]{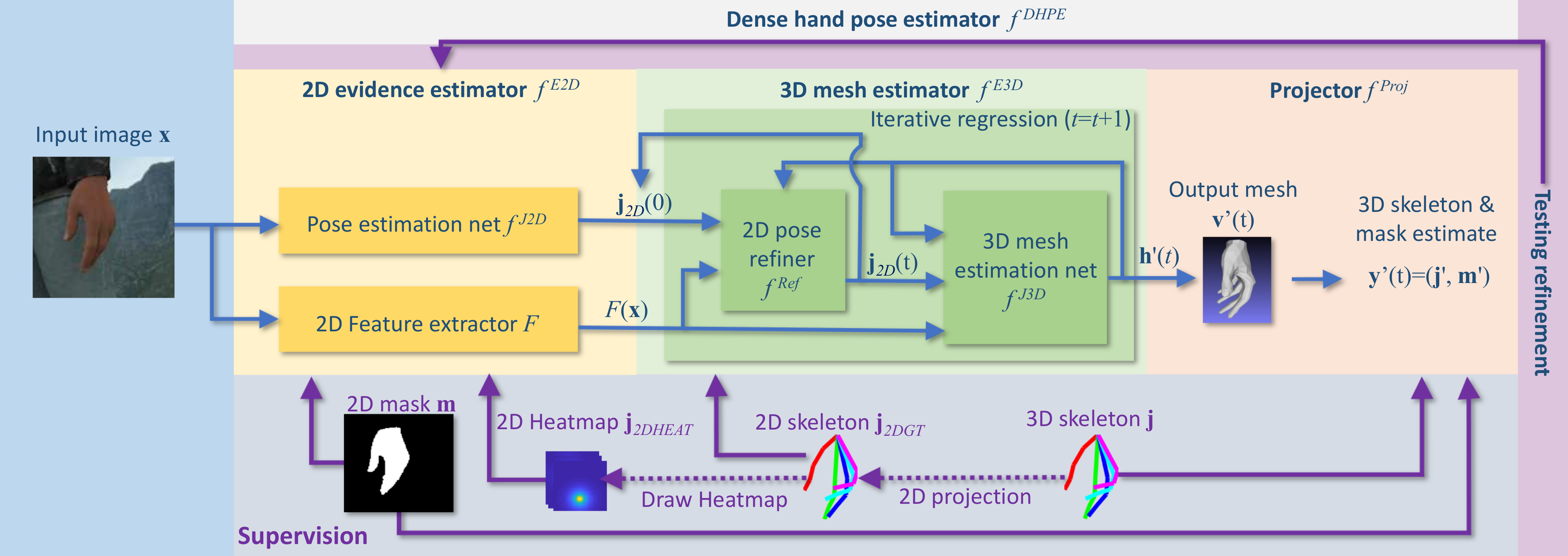}
\caption{Schematic diagram of the proposed dense hand pose estimation (DHPE) framework. Our DHPE receives an input RGB image $\mbx$ and estimates the corresponding hand shape and pose as parameters $\mbh$ of the MANO hand model. Training DHPE is guided via 1) an additional projector $f^{\text{\emph{Proj}}}$ that enables us to provide supervision via 3D skeletons $\mbj$ and foreground segmentation masks $\mbm$; 2) decomposing the DHPE into the 2D evidence estimator $f^{\text{\emph{E2D}}}$ and 3D mesh estimator $f^{\text{\emph{E3D}}}$ which stratifies the training via the intermediate 2D feature estimation step. At testing, once the output mesh parameter $\mbh'$ is estimated, it is iteratively refined via enforcing its consistency over intermediate 2D evidences $F(\mbx)$ and  
$\mbj_{2D}$.}
\label{fig:arch}
\end{figure*}

\emph{Estimating the 2D skeletal from RGB.} 
Recovering 2D skeletal representations from RGB images has been greatly improved~\cite{hand_cvpr_2017_1}. Compared to the depth domain, real RGB data is scarce and automatic pose annotation of RGB images still remains challenging. Simon~\etal achieved promising results by mixing real data with a large amount of synthetic data~\cite{hand_cvpr_2017_1}. Further accuracy improvement has been obtained by the automatic annotation using label consistency in a multiple camera studio~\cite{multiviewsystem}.

\emph{Recovering the 3D skeletal from RGB.}
The problem has inherent uncertainties rising from 2D-to-3D mapping~\cite{eccv2018_rgbhand2,eccv2018_rgbhand,iccv_2017_zimmerman}. The earlier work~\cite{iccv_2017_zimmerman} attempts to learn the direct mapping from RGB images to 3D skeletons. Recent methods~\cite{eccv2018_rgbhand,eccv2018_rgbhand2} have shown the state-of-the-art accuracy by implicitly reconstructing depth images i.e. 2.5D representations, and estimating the 3D skeletal based on them. We instead use a 3D mesh model in a hybrid way to tackle the problem.

\emph{3D representations for hand pose estimation.} Different from human face or body, hand poses exhibit a much wider scope of viewpoints, given an articulation and shape. It is hard to define a canonical viewpoint (e.g. a frontal view). To relieve the issue, intermediate 3D representations such as projected point clouds~\cite{hand_cvpr_2016_1}, D-TSDF~\cite{hand_cvpr_2017_4}, voxels~\cite{hand_cvpr18_v2v} or point sets~\cite{cvpr2018handpointnet} have been proposed. Though these approaches have shown promising results, they are limited such that they cannot re-generate full 3D information.
\vspace{1mm}

\noindent\textbf{3D recovery in other domains. }
3D reconstruction has a long history in the field. Some latest approaches using neural networks are discussed below.

\emph{3D voxel reconstruction.}
Voxel representations are useful to describe 3D aspects of objects \cite{3d_nips_2016,cvpr_2018_multimaskvoxel}. Yan~\etal~\cite{3d_nips_2016} proposed a differentiable way to reconstruct voxels given only 2D inputs. Also, Tulsiani~\etal~\cite{cvpr_2018_multimaskvoxel} proposed a method to use multiple images caputred at different viewpoints to accurately reconstruct voxels. However, the voxel representation is sparse compared to the mesh counterpart. To relieve the issue, a hierarchical method was proposed by Riegler~\etal~\cite{cvpr2017_octnet}. 

\emph{3D mesh reconstruction.}
Recently, G{\"u}ler~\etal~\cite{cvpr_2018_densepose} proposed a way to reconstruct human bodies from  RGB inputs. This method requires full 3D supervisions from lots of pairs of 2D images and 3D mesh annotations. Differentiable renderers~\cite{eccv_2014_opendr,cvpr_2018_neuralrenderer} have been proposed such that 2D silhouettes, depth maps and even textured RGB maps are obtained by projecting 3D models to image planes in a differentiable way. With the use of this renderer, it  becomes much easier to reconstruct 3D meshes~\cite{eccv_2018_mesh}, not needing full 3D supervisions. End-to-end CNN methods with the SMPL model~\cite{SMPL:2015} have been proposed for recovering 3D meshes of human bodies~\cite{nips_2017_motioncapture,e2eshapepose,bodyrecon_cvpr18}. Hands are, however, different from human bodies: hands exhibit much wider viewpoint variations (both egocentric/third-person views), highly articulated and frequently occluded joints. Also, there are relatively fewer full 3D annotations of RGB images available. These domain differences make the na\"{i}ve application of the existing methods to this domain non-optimal.

\section{Proposed dense hand pose estimator}

Our goal is to construct a \emph{hand mesh estimator} (HME) $f^{\text{\emph{HME}}}:\calX\to\calV$ that maps an input RGB image $\mbx\in \calX$ to the corresponding estimated 3D hand mesh $\mbv\in \calV$. Each input is given as an RGB pixel array of size 224$\times$224 while an output mesh corresponds to 3D positions of 778 vertices encoding 1,538 triangular faces. Instead of directly generating a mesh $\mbv$ as a $778\times 3$-dimensional raw vector, our HME estimates a 63-dimensional parameter vector $\mbh$ that represents $\mbv$ by adopting the MANO 3D hand mesh model~\cite{mano} (Sec.~\ref{s:3dmeshestimator}). Once a hand mesh $\mbv$ (or equivalently, its parameter $\mbh$) is estimated, the corresponding skeletal pose $\mbj$ consisting of 21 3D joint positions can be recovered via a 3D skeleton regressor $f^{\text{\emph{Reg}}}$ discussed shortly.

Learning the HME can be cast into a standard multivariate regression problem if a training database of pairs of input images and the corresponding ground-truth meshes are available. However, we are not aware of any existing database that provides such pairs. Inspired by the success of existing human body reconstruction work~\cite{nips_2017_motioncapture,e2eshapepose,bodyrecon_cvpr18}, we take an indirect approach by learning a \emph{dense hand pose estimator (DHPE)} which combines the HME and a new \emph{projection operator}. The projection operator consists of a 3D \emph{skeleton regressor} and a \emph{renderer} which respectively recover a 3D skeletal pose ($\mbj\in\calJ$) and a 2D foreground hand segmentation mask ($\mbm\in\calM$) from a hand mesh $\mbv$. Table~\ref{t:dhpenotation} shows our notation, and Eq.~\ref{e:dhpedecmoposition} and Fig.~\ref{fig:arch} summarize the decomposition of DHPE:
\begin{align}
\label{e:dhpedecmoposition}
\calX\underbrace{\overbrace{\xrightarrow{f^{\text{\emph{E2D}}}}\calZ\xrightarrow{f^{\text{\emph{E3D}}}}}^{f^{\text{\emph{HME}}}=f^{\text{\emph{E3D}}}\circ f^{\text{\emph{E2D}}}}\calV\xrightarrow{f^{\text{\emph{Proj}}}=(f^{\text{\emph{Reg}}},f^{\text{\emph{Ren}}})}}_{f^{\text{\emph{DHPE}}}=f^{\text{\emph{Proj}}}\circ f^{\text{\emph{HME}}}}\calY=(\calJ,\calM).
\end{align}

\begin{table}[t]
\caption{Notational summary}
\label{t:dhpenotation}
\resizebox{\linewidth}{!}{
\begin{tabular}{|c|l|}
\hline
$\mby$& $\mby=(\mbj,\mbm)\in \calY=(\calJ,\calM)\subset \R^{(21\times 3)\times (224\times 224)}$\\
\hline
$f^{\text{\emph{DHPE}}}$& dense hand pose estimator ($f^{\text{\emph{DHPE}}}=f^{\text{\emph{Proj}}}\circ f^{\text{\emph{HME}}}$)\\
\hline
$f^{\text{\emph{Proj}}}$ & projection operator ($f^{\text{\emph{Proj}}}=[f^{\text{\emph{Reg}}},f^{\text{\emph{Ren}}}]$)\\
\hline
$f^{\text{\emph{Reg}}}$& 3D skeleton regressor\\
\hline
$f^{\text{\emph{Ren}}}$& renderer\\
\hline
\end{tabular}
}
\vspace{-0.5cm}
\end{table}

Extending the skeleton regressor provided by the MANO model~\cite{mano}, our skeleton regressor $f^{\text{\emph{Reg}}}$ maps a hand mesh $\mbv$ to its skeleton $\mbj$ consisting of 21 3D joint positions. It is a linear regressor implemented as three matrices of size $778\times 21$ (see Sec.~2.1 of the supplemental for details).

Our renderer $f^{\text{\emph{Ren}}}$ generates the foreground hand mask $\mbm$ by simulating the camera view of $\mbx$. We adopt the differentiable neural renderer proposed by Kato~\etal~\cite{cvpr_2018_neuralrenderer}.

By construction, the projection operator respects the underlying camera (via $f^{\text{\emph{Ren}}}$) and hand shape (via $f^{\text{\emph{Reg}}}$) geometry and it is held fixed throughout the entire training process. This facilitates the training of $f^{\text{\emph{HME}}}$ indirectly via training $f^{\text{\emph{DHPE}}}$. However, even under this setting, the problem still remains challenging as estimating a 3D mesh given an RGB image is a seriously ill-posed problem. Adopting recent human body pose estimation approaches~\cite{bodyrecon_cvpr18,nips_2017_motioncapture}, we further stratify learning of $f^{\text{\emph{DHPE}}}:\calX\to\calY$ by decomposing $f^{\text{\emph{HME}}}:\calX\to\calV$ into a 2D evidence estimator $f^{\text{\emph{E2D}}}:\calX\to\calZ$ and a 3D mesh estimator $f^{\text{\emph{E3D}}}:\calZ\to\calV$.  Our 2D evidence $\mbz\in Z$ consists of a 42-dimensional 2D skeletal joint position vector $\mbj_{2D}$ (21 positions $\times$ 2; as in~\cite{eccv2018_rgbhand,eccv2018_rgbhand2}) and a 2,048-dimensional 2D feature vector $F(\mbx)$ (Eq.~\ref{loss_feat}). The remainder of this section provides details of these two estimators.

\subsection{2D evidence estimator $f^{\text{\emph{E2D}}}=(F,f^{\text{\emph{J2D}}})$}
\label{s:2devidence}
Silhouettes (or foreground masks) and 2D skeletons have been widely used as the mid-level cues for estimating 3D body meshes~\cite{eccv_2018_mesh,bodyrecon_cvpr18,nips_2017_motioncapture}. However, for hands, accurately estimating foreground masks from RGB images is challenging due to cluttered backgrounds~\cite{cvpr2018_handseg, iccv_2017_zimmerman}. We observed that na\"{i}vely applying the state-of-the-art foreground segmentation algorithms (\eg~\cite{iccv_2017_zimmerman}) often misses fine details, especially along the narrow finger regions (see Fig.~\ref{fig:segresult} for examples) and this can significantly degrade the performance of the subsequent mesh estimation step. 
We bypass this challenge by learning instead, a 2D foreground feature extractor $F$: $F(\mbx)$ encapsulates the textural and shape information of the foreground regions in $\mbx$.

\begin{figure}[!t]
\centering
\captionsetup[subfigure]{labelformat=empty}
\subfloat[(a)]{\includegraphics[width=0.23\linewidth]{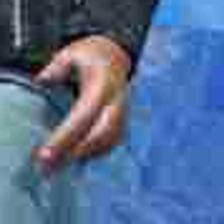}}
\subfloat[(b)]{\includegraphics[width=0.23\linewidth]{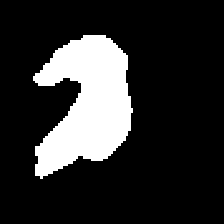}}
\subfloat[(c)]{\includegraphics[width=0.23\linewidth]{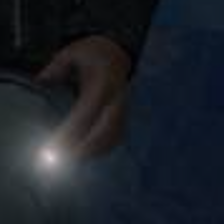}}
\subfloat[(d)]{\includegraphics[width=0.23\linewidth]{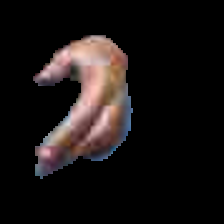}}
\vspace{-2mm}
\caption{A 2D evidence estimation example. (a) input image $\mbx$, (b) ground-truth 2D segmentation mask $\mbm$ of $\mbx$, (c) 2D skeletal position heat map of the finger tip of middle finger overlaid on $\mbx$, and (d) masked image $\mbx\odot \mbm$.}
\label{fig:second}
\end{figure}

\vspace{1mm}
\noindent\textbf{Our 2D feature extractor $F$} is trained to focus on the foreground regions by minimizing the deviation between the features extracted from the entire image $\mbx$ and the foreground region $\mbx \odot \mbm$ extracted via the ground-truth mask $\mbm$ (see Fig.~\ref{fig:second}d): The training loss (per data point) for $F$ is given as
\begin{align}
L_{\text{\emph{Feat}}}(F) = \|F(\mbx)-F(\mbx \odot \mbm)\|^2_2,
\label{loss_feat}
\end{align}
where $\odot$ denotes element-wise multiplication. $F$ employs the ResNet-50~\cite{ResNet50} architecture whose output is a 2,048-dimensional vector.

\vspace{1mm}
\noindent\textbf{Estimating 2D skeletal joints from an RGB image} has been well studied. Similarly to \cite{eccv2018_rgbhand,eccv2018_rgbhand2}, we embed the state-of-the-art 2D pose estimation network $f^{\text{\emph{J2D}}}$~\cite{cpm_cvpr2016,iccv_2017_zimmerman} into our framework. The initial weights provided by the authors of ~\cite{iccv_2017_zimmerman} are refined based on the 2D joint estimation loss:

\begin{align}
L_{\text{\emph{J2D}}}(f^{\text{\emph{J2D}}}) = \|f^{\text{\emph{J2D}}}(\mbx)-\mbj_{\text{\emph{2DHeat}}}\|^2_2.
\label{loss_j2d}
\end{align}
The output of $f^{\text{\emph{J2D}}}$ is a matrix of size 21$\times$32$\times$32 encoding 21 heat maps (per joint) of size 32$\times$32. $\mbj_{\text{\emph{2DHeat}}}$ is the ground-truth heat map. Given the estimated heat maps, 2D skeleton joints $\mbj_\text{\emph{2D}}$ are extracted by finding the maximum for each joint. Figure~\ref{fig:second}(b) shows an example of 2D evidence estimation. 

\subsection{3D mesh estimator $f^{\text{\emph{E3D}}}=(f^{\text{\emph{J3D}}},f^{\text{\emph{Ref}}})$}
\label{s:3dmeshestimator}
Taking the 2D evidence $\mbz$ ($2,048$-dimensional feature vector $F(\mbx)$ and $21\times 3$-dimensional 2D skeleton $\mbj_\text{\emph{2D}}$) as input, the 3D mesh estimation network $f^{\text{\emph{J3D}}}$ constructs parameters of a deformable hand mesh model and a camera model. We use the MANO model representing a hand mesh based on 45-dimensional pose parameters $\mbp$ and 10-dimensional shape parameters $\mbs$~\cite{mano}. The original MANO framework uses only 6-dimensional (principal component analysis) PCA subspace of $\mbp$ for computational efficiency. However, we empirically observed that to cover a variety of hand poses, all 45-dimensional features are required: We use the linear blend skinning formulation as in the SMPL model~\cite{SMPL:2015}. 

Given the MANO parameters $\mbp$ and $\mbs$, a mesh $\mbv$ is synthesized by conditioning them on the hypothesized camera $\mbc$, which comprises of the 3D rotation (in quaternion space) $\mbc_q\in \R^4$, scale $\mbc_s\in \R$, and translation $\mbc_t\in \R^3$: An initial mesh is constructed by combining MANO's PCA basis vectors using the shape parameter $\mbs$, rotating the bones according to the pose parameter $\mbp$, and deforming the resulting surface via linear blend skinning. The final mesh $\mbv$ is then obtained by globally rotating, scaling, and translating the initial mesh according to $\mbc_q$, $\mbc_s$, and $\mbc_t$, respectively. The entire mesh generation process is differentiable. Under this model, our 3D mesh estimation network $f^{\text{\emph{J3D}}}$ is implemented as a single fully-connected layer of 2153$\times$63 weights and it estimates a 63-dimensional mesh parameter:
\begin{align}
\mbh = [\mbp, \mbs, \mbc_q, \mbc_s, \mbc_t]^\top.
\label{output_p}
\end{align}

\noindent\textbf{Iterative mesh refinement via back-projection.} Adopting Kanazawa~\etal's approach~\cite{e2eshapepose}, instead of estimating $\mbh$ directly, we iteratively refine the initial mesh estimate $\mbh(0)$ by recursively performing regression on the parameter offset $\Delta\mbh$: At iteration $t$, $f^{\text{\emph{J3D}}}$ takes $\mbz$ and the current mesh estimate $\mbh(t)$, and generates a new offset $\Delta\mbh(t)$:
\begin{align}
\mbh(t+1)=\mbh(t)+\Delta\mbh(t).
\label{e:iterativeregression}
\end{align}
At $t=0$, $\mbh(0)$ as an input to $f^{\text{\emph{J3D}}}$, is constructed as a vector of zeros except for the entries corresponding to the pose parameter $\mbp$ which is set as the mean pose of the MANO model. We fix the number of iterations at 3 as more iterations did not show any noticeable improvements in preliminary experiments.

In \cite{e2eshapepose}, each offset prediction $\Delta\mbh(t)$ is built based only on 2D features $F(\mbx)$. Inspired by the success of \cite{eccv2018_rgbhand,eccv2018_rgbhand2}, we additionally use 2D skeletal joint $\mbj'_{2D}$ as input. However, as an estimate, $\mbj'_{2D}(0)$ (obtained at iteration $0$) is inaccurate, causing errors in the resulting $\Delta\mbh(t)$ estimate. Then, these errors accumulate over iterations (Eq.~\ref{e:iterativeregression}) lessening the benefit of the entire iterative estimation process. We address this by an additional 2D pose refiner $f^{\text{\emph{Ref}}}$ which iteratively refines the 2D joint estimation $\mbj'_{2D}(t)$ via \emph{back-projecting} the estimated mesh $\mbh'(t)$: At $t$, $f^{\text{\emph{Ref}}}$ receives the estimated mesh parameter $\mbh'(t)$, 2D feature $F(\mbx)$, 3D skeleton $f^{\text{\emph{Reg}}}(\mbv')$, and $\mbj'_{2D}(t)$, and generates a refined estimate $\mbj'_{2D}(t+1)$. The 2D pose refiner $f^{\text{\emph{Ref}}}$ is implemented as a single fully connected layer of size 2216$\times$42 and it is trained based on the ground-truth 2D skeletons ($\mbj_{\text{\emph{2DGT}}}$):

\begin{eqnarray}
L_{\text{\emph{Ref}}} = \left\|f^{\text{\emph{Ref}}}\left([\mbj'_{2D}(t),F(\mbx),\mbh'(t),f^{\text{\emph{Reg}}}(\mbv')]\right)-\mbj_{\text{\emph{2DGT}}}\right\|^2_2.
\label{loss_j2drefine}
\end{eqnarray}

In the experiments, we demonstrate that 1) using both 2D features and 2D skeletons (Fig.~\ref{fig:results}(e): `Ours (w/o Test. ref. and Refiner $f^{\text{\emph{Ref}}}$)') improves performance over Kanazawa~\etal's 2D feature-based framework~\cite{e2eshapepose} (Fig.~\ref{fig:results}(e): `Ours (w/o Test. ref., $f^{\text{\emph{Ref}}}$ and 2D losses ($L_{Feat}$, $L_{J2D}$)'); 2) explicitly building the refiner $f^{\text{\emph{Ref}}}$ under the auxiliary 2D joint supervision (Fig.~\ref{fig:results}(e): `Ours (w/o Test. ref.)') further significantly improves performance. More importantly, the refiner takes a crucial role in our testing refinement step (Sec.~\ref{s:testingrefinement}) which improves a once predicted 3D mesh $\mbv'$ by enforcing its consistency with the corresponding 2D joint evidence $\mbj'_\text{\emph{2D}}(t)$ (as well as other intermediate results) throughout the iteration (Eq.~\ref{refine_test}).

\ignore{In the experiments, we demonstrate that 1) using \seung{the 2D losses in Eqs.~\ref{loss_feat},~\ref{loss_j2d}} (Fig.~\ref{fig:results}(e): `Ours (w/o Test. ref. and Refiner $f^{\text{\emph{Ref}}}$)' improves performance over Kanazawa~\etal's 2D feature-based framework~\cite{e2eshapepose} (Fig.~\ref{fig:results}(e): `Ours (w/o Test. ref., \seung{$f^{\text{\emph{Ref}}}$} and 2D losses \seung{($L_{Feat}$, $L_{J2D}$)}'); 2) explicitly building the refiner $f^{\text{\emph{Ref}}}$ under the auxiliary 2D joint supervision further significantly improves performance. More importantly, \seung{the refiner $f^{\text{\emph{Ref}}}$} takes a crucial role in our testing refinement step (Sec.~\ref{s:testingrefinement}) which improves a once predicted 3D mesh $\mbv'$ by enforcing its consistency with the corresponding 2D joint evidence $\mbj'_\text{\emph{2D}}(t)$ (as well as other intermediate results) throughout the iteration (Eq.~\ref{e:iterativeregression}). We fix the number of iterations to $3$ as a larger number of iterations did not show  noticeable improvements. }

\subsection{Joint training}
\label{s:trainingprocess}
Training the 2D evidence estimator $f^{\text{\emph{E2D}}}:\calX\to\calZ$ and 3D mesh estimator $f^{\text{\emph{E3D}}}:\calZ\to\calV$  
given fixed projection operator $f^{\text{\emph{Proj}}}:\calV\to\calY$ is performed based on a training set consisting of input images, and the corresponding 3D skeleton joints and 2D segmentation masks $D=\{(\mbx_i,\mby_i\}_{i=1}^l\subset \calX\times \calY$, $\mby_i=(\mathbf{j}_i,\mathbf{m}_i)$. Since all component functions of $f^{\text{\emph{E2D}}}$, $f^{\text{\emph{E3D}}}$, and $f^{\text{\emph{Proj}}}$ are differentiable with respect to the weights of $F$ and $f^{\text{\emph{E3D}}}$, they can be optimized based on on standard gradient descent-type algorithm: Our overall loss (per training instance $\mbx_i,\mby_i$) is given as (see Eqs.~\ref{loss_feat} and \ref{loss_j2drefine}):
\begin{align}
L(f^{\text{\emph{E3D}}},F) &= L_\text{\emph{Art}}(f^{\text{\emph{E3D}}},F)+ L_\text{\emph{Lap}}(f^{\text{\emph{E3D}}},F)+L_{\text{\emph{Feat}}}(F)\nonumber\\
&+ \lambda L_\text{\emph{Sh}}(f^{\text{\emph{J3D}}},F)+L_{\text{\emph{Ref}}}(f^{\text{\emph{J3D}}},F).
\label{loss_3d}
\end{align}

\noindent\textbf{The articulation loss $L_\text{\emph{Art}}$} measures the deviation between the skeleton estimated from $\mbx_i$ and its ground-truth $\mbj_i$:
\begin{eqnarray}
L_\text{\emph{Art}} =\|[f^{\text{\emph{DHPE}}}(\mbx_i)]_{\calJ}-\widehat{\mbj_i}\|^2_2,
\label{loss_art}
\end{eqnarray}
where $[\mby]_{\calJ}$ extracts the $\mbj$-component of $\mby=(\mbj,\mbm)$ and $\widehat{\mbj}$ spatially normalizes $\mbj$ similarly to 
\cite{eccv2018_rgbhand2,iccv_2017_zimmerman}: 
First, the center of each skeleton is moved to the corresponding middle finger's MCP position. Then each axis is normalized to a unit interval $[0,1]$: The $x,y-$coordinate values are divided by $g$ (=1.5 times the maximum of height and width of the tight 2D hand bounding box). The $z-$axis value is divided by $(z_{\text{\emph{Root}}}\times g)/\mbc_f$ where $z_{\text{\emph{Root}}}$ is the depth value of the middle finger's MCP joint and $\mbc_f$ is the focal length of the camera. At testing, once normalized skeletons are estimated, they are inversely normalized to the original scale. The accompanying supplemental provides details of this inverse normalization step.

\vspace{1mm}
\noindent\textbf{Our shape loss $L_\text{\emph{Sh}}$} facilitates the recovery of hand shapes as observed indirectly via projected 2D segmentation masks:
\begin{eqnarray}
L_\text{\emph{Sh}} = \left\|[f^{\text{\emph{DHPE}}}(\mbx_i)]_{\calM}-\mbm_i\right\|^2_2,
\label{loss_sh}
\end{eqnarray}
where $[\mby]_{\calM}$ extracts the $\mbm$-component of $\mby=(\mbj,\mbm)$. 

\vspace{1mm}
\noindent\textbf{The Laplacian regularizer $L_\text{\emph{Lap}}$} enforces spatial smoothness in the mesh $\mbv$. This helps avoid generating implausible hand meshes as suggested by Kanazawa~\etal~\cite{eccv_2018_mesh}.

\vspace{1mm}
\noindent\textbf{Hierarchical recovery of articulation and shapes.} We observed that na\"{i}vely minimizing the overall loss $L$ with a constant shape loss weight $\lambda$ (Eq.~\ref{loss_3d}) tends to impede convergence during training (Fig.~\ref{fig:results}(f)): Our algorithm simultaneously optimizes the mesh ($\mbv$) and camera ($\mbc=\{\mbc_q,\mbc_s,\mbc_t\}$) parameters which over-parameterize the rendered 2D view, \eg the effect of scaling $\mbv$ itself can be offset by inversely scaling $\mbc_s$. This often hinders the alignment of $\mbv$ with the ground-truth 2D mask and thereby rendering the network inappropriately update the mesh parameters. Therefore, we let the shape loss $L_\text{\emph{Sh}}$ take effect only when the articulation loss $L_\text{\emph{Art}}$ becomes sufficiently small: $\lambda$ is set per data instance based on the $L_\text{\emph{Art}}$-value:
\begin{eqnarray}
\lambda = \left\{ \begin{array}{ll}
1 & \textrm{if $\left\|\left[[f^{\text{\emph{DHPE}}}(\mbx_i)]_\mbj\right]_{2D} -[\mbj_i]_{2D}\right\|^2_2<\tau$}\\
0 & \textrm{otherwise},
\end{array} \right.
\label{eq:hierloss}
\end{eqnarray}
where $[\mbj]_{2D}$ projects 3D joint coordinates $\mbj$ to 2D view based on $\mbc$ (Eq.~\ref{output_p}). The threshold $\tau$ is empirically set to $15$ pixels. Initially, with zero $\lambda$, $L_\text{\emph{Art}}$ dominates in $L$, which helps globally align the estimated meshes with 2D ground-truth evidence. As the training progresses, the role of $L_\text{\emph{Sh}}$ becomes more important ($\lambda=1$) contributing to recovering the detailed shapes.

As the generation of the 2D skeleton $\mbj_{\text{\emph{2D}}}$ from an estimated heat-map $f^{\text{\emph{J2D}}}(\mbx)$ is not differentiable (see~Eq.~\ref{loss_j2d}), our 2D pose estimation network $f^{\text{\emph{J2D}}}$ cannot be trained based on $L$. Thus we train it  in parallel using $L_{\text{\emph{J2D}}}$ (Eq.~\ref{loss_j2d}). For both cases, we use the standard Adam optimizer with the learning rate $\gamma$ set at $10^{-3}$.

\subsection{Testing refinement} 
\label{s:testingrefinement}
To facilitate the training of the HME, we constructed an auxiliary DHPE that decomposes into three component functions: $f^{\text{\emph{Proj}}}$, $f^{\text{\emph{E2D}}}$, and $f^{\text{\emph{E3D}}}$ (see Eq.~\ref{e:dhpedecmoposition}). An important benefit of this step-wise estimation approach is that it enables us to check and improve once predicted output mesh by comparing it with the intermediate results: For testing, the underlying mesh $\mbv$ of a given test image $\mbx$ can be first estimated by applying $f^{\text{\emph{HME}}}=f^{\text{\emph{E3D}}}\circ f^{\text{\emph{E2D}}}:\calX\to\calV$. If the resulting prediction $\mbv'$ (equivalently, $\mbh'$) is accurate, it must be in accordance with the intermediate results $F(\mbx)$ and $\mbj_\text{\emph{J2D}}'$ generated from $\mbx$. Checking and further enforcing this consistency can be facilitated by noting that our loss function $L$ and its components are differentiable with respect to the mesh parameter $\mbh$. By reinterpreting $L$ as a smooth function of $\mbh$ given fixed $f^{\text{\emph{Proj}}}$, $f^{\text{\emph{E2D}}}$, and $f^{\text{\emph{E3D}}}$, we can refine the initial prediction $\mbh'(0)$ by enforcing such consistency:
\begin{align}
\mbh'(t+1) &= \mbh'(t)-\gamma\cdot\nabla_{\mbh}\Big(\left\|\left[[f^{\text{\emph{DHPE}}}(\mbx)]_{\calJ}\right]_{XY}-{\mbj'}_\text{\emph{J2D}}\right\|^2_2\nonumber\\
&+\lambda\left\|F(\mbx)-F(f^{\text{\emph{Ren}}}(\mbv')\odot \mbx)\right\|^2_2+ L_\text{\emph{Lap}}\Big),
\label{refine_test}
\end{align}

where $[\mbj]_{XY}$ extracts the $x,y-$coordinate values of skeleton joints from $\mbj$. Note that in the first gradient term, we use the 2D skeleton ${\mbj'}_\text{\emph{2D}}$ since the ground-truth 3D skeleton $\widehat{\mbj}$ is not available at testing. This step benefits from explicitly building the 2D pose refiner $f^{\text{\emph{Ref}}}$ that improves the once estimated 2D joint ${\mbj'}_\text{\emph{2D}}$ during iterative mesh estimation process (Eq.~\ref{e:iterativeregression}). Also, for the second, shape loss term of the gradient, since the ground-truth segmentation mask $\mbm$ is not available, we use the segmentation mask rendered via $f^{\text{\emph{Ren}}}$ based on the estimated mesh $\mbv'$ enforcing self-consistency. The number of iterations in Eq.~\ref{refine_test} is fixed at 50. Our testing refinement step takes 250ms (5ms per iteration $\times$ 50 iterations) in addition to the initial regression step which takes 100ms. Figure~\ref{fig:results}(e) shows the performance variation with varying number of iterations. 

\subsection{Self-supervised data augmentation}
\label{s:dataaugmentation}
An important advantage of incorporating a generative mesh model (\ie MANO) into the training process of the HME (via $\mbh$; see Eq.~\ref{output_p}) is that it can synthesize new data as guided by (and subsequently, guiding) HME. The MANO model provides explicit control over the shape of synthesized 3D hand mesh.\footnote{While it is also possible to generate new poses, we do not explore this possibility since we observed that it often leads to implausible hand poses.} Combining this with a camera model, we can generate pairs of 3D meshes, and the corresponding rendered 2D masks and RGB images.

To render RGBs, we adopt the neural texture renderer $f^\text{\emph{TRen}}$ proposed by Kato~\etal~\cite{cvpr_2018_neuralrenderer}: During training, once a seed mesh $\mbv^S$ is predicted, the corresponding camera and shape parameters $\mbh^S$ are changed to generate new meshes $\{\mbv^\text{\emph{N}}_j\}$: The shape parameter $\mbs$ is sampled uniformly from the interval covering three times the standard deviation per dimension. For camera perspectives, the rotation matrix along each of $x, y, z-$ axes are sampled uniformly on $[0, 2\pi]$. Once the foreground hand region is rendered via $f^\text{\emph{TRen}}$, it is placed on random backgrounds obtained from the NYU depth database~\cite{nyubg}.

For each new mesh $\mbv^\text{\emph{N}}$, we generate a triplet $\left(f^{\text{\emph{TRen}}}(\mbv^\text{\emph{N}}), f^{\text{\emph{Ren}}}(\mbv^\text{\emph{N}}),f^{\text{\emph{Reg}}}(\mbv^\text{\emph{N}})\right)$ constituting a new training instance for the DHPE. We empirically observed that when the training of the DHPE reaches $20$ epochs, it tends to generate seed meshes $\{\mbv^S_i\}$ which faithfully represent realistic hand shapes (even though they might not accurately match the corresponding input images $\{\mbx_i\}$). Therefore, we initiate the augmentation process after the first $20$ training epochs. For each mini-batch, three new data instances are generated per seed prediction $\mbv^S$ gradually enlarging the entire training set (see supplemental for examples). A similar self-supervised data augmentation approach has been adopted for facial shape estimation~\cite{kim2018inverse}.

\begin{figure}[t]
\centering
\captionsetup[subfigure]{labelformat=empty}

\subfloat{\includegraphics[width=0.245\linewidth]{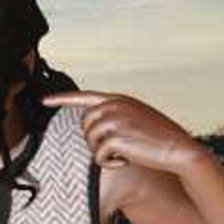}}
\subfloat{\includegraphics[width=0.245\linewidth]{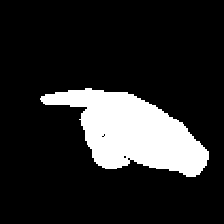}}
\subfloat{\includegraphics[width=0.245\linewidth]{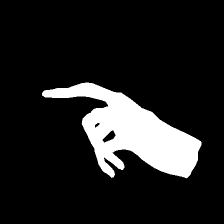}}
\subfloat{\includegraphics[width=0.245\linewidth]{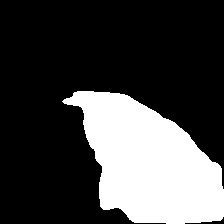}}\\
\vspace{-0.375cm}
\subfloat[]{\includegraphics[width=0.245\linewidth]{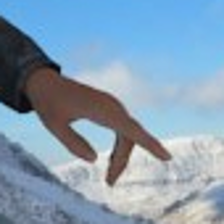}}
\subfloat[]{\includegraphics[width=0.245\linewidth]{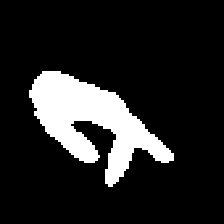}}
\subfloat[]{\includegraphics[width=0.245\linewidth]{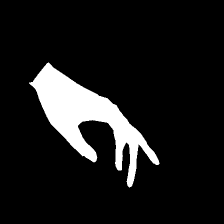}}
\subfloat[]{\includegraphics[width=0.245\linewidth]{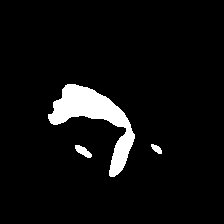}}
\vspace{-4mm}
\caption{Hand segmentation examples. Left to right: input images, ground-truth masks, our results, the results of the state-of-the-art hand segmentation algorithm~\cite{cvpr2018_handseg}.}
\label{fig:segresult}
\end{figure}

\section{Experiments}
\begin{figure*}[!t]
\centering
\captionsetup[subfigure]{labelformat=empty}
\subfloat[]{\includegraphics[width=0.082\linewidth]{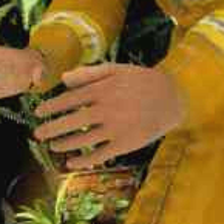}}
\subfloat[]{\includegraphics[width=0.082\linewidth]{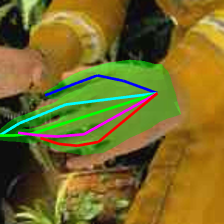}}
\subfloat[]{\includegraphics[width=0.082\linewidth]{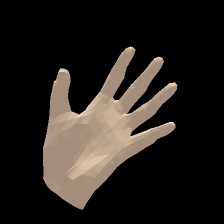}}
\subfloat[]{\includegraphics[width=0.082\linewidth]{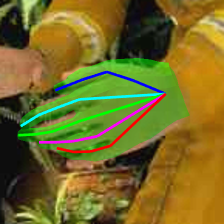}}
\subfloat[]{\includegraphics[width=0.082\linewidth]{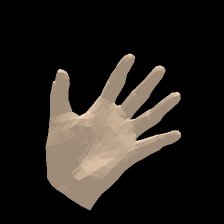}}
\hspace{+0.03cm}
\subfloat[]{\includegraphics[width=0.082\linewidth]{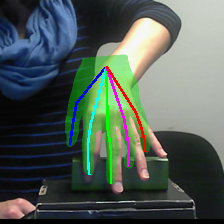}}
\subfloat[(ii)]{\includegraphics[width=0.082\linewidth]{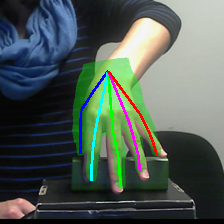}}
\subfloat[]{\includegraphics[width=0.082\linewidth]{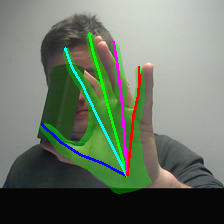}}
\subfloat[]{\includegraphics[width=0.082\linewidth]{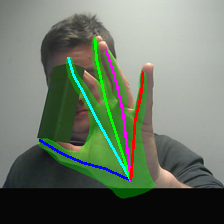}}
\subfloat[]{\includegraphics[width=0.082\linewidth]{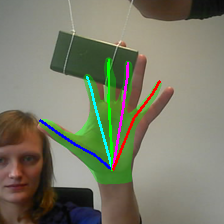}}
\hspace{+0.03cm}
\subfloat[]{\includegraphics[width=0.082\linewidth]{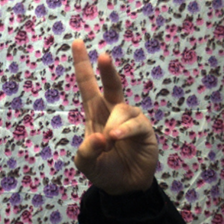}}
\subfloat[]{\includegraphics[width=0.082\linewidth]{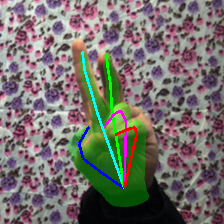}}
\\
\vspace{-0.86cm}
\subfloat[]{\includegraphics[width=0.082\linewidth]{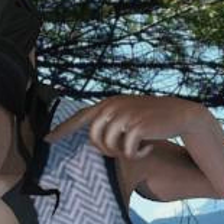}}
\subfloat[]{\includegraphics[width=0.082\linewidth]{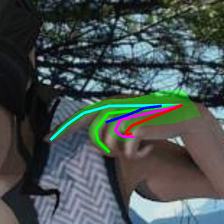}}
\subfloat[]{\includegraphics[width=0.082\linewidth]{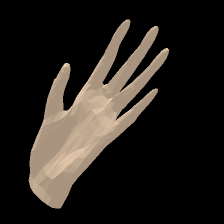}}
\subfloat[]{\includegraphics[width=0.082\linewidth]{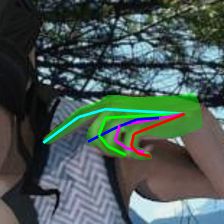}}
\subfloat[]{\includegraphics[width=0.082\linewidth]{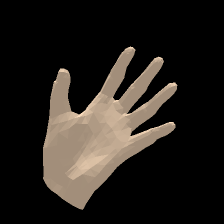}}
\hspace{+0.03cm}
\subfloat[]{\includegraphics[width=0.082\linewidth]{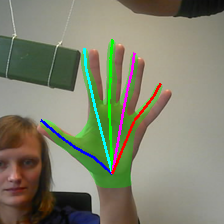}}
\subfloat[a]{\includegraphics[width=0.082\linewidth]{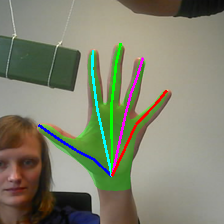}}
\subfloat[]{\includegraphics[width=0.082\linewidth]{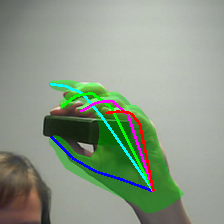}}
\subfloat[]{\includegraphics[width=0.082\linewidth]{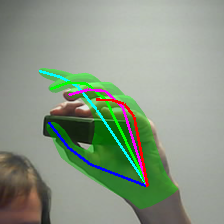}}
\subfloat[]{\includegraphics[width=0.082\linewidth]{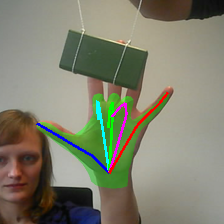}}
\hspace{+0.03cm}
\subfloat[]{\includegraphics[width=0.082\linewidth]{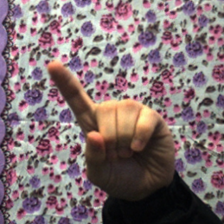}}
\subfloat[]{\includegraphics[width=0.082\linewidth]{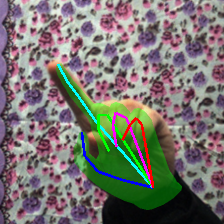}}
\\
\vspace{-1.2665cm}
\hspace{-0.165cm}
\subfloat[(\emph{RHD})]{
\subfloat[(a)]{\includegraphics[width=0.082\linewidth]{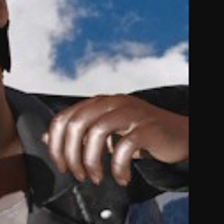}}
\subfloat[(b)]{\includegraphics[width=0.082\linewidth]{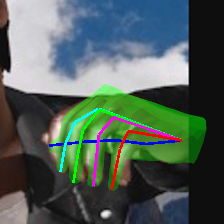}}
\subfloat[(c)]{\includegraphics[width=0.082\linewidth]{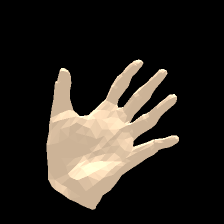}}
\subfloat[(d)]{\includegraphics[width=0.082\linewidth]{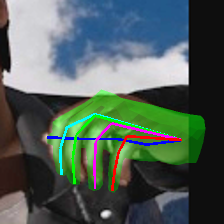}}}
\subfloat[(e)]{\includegraphics[width=0.082\linewidth]{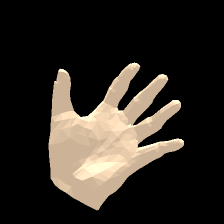}}
\hspace{-0.06cm}
\subfloat[(\emph{DO})]{
\subfloat[(a)]{\includegraphics[width=0.082\linewidth]{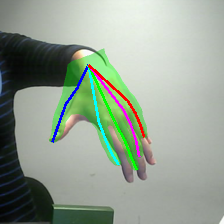}}
\subfloat[(b)]{\includegraphics[width=0.082\linewidth]{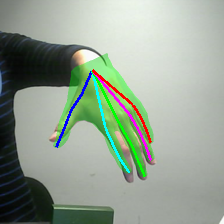}}
\subfloat[(c)]{\includegraphics[width=0.082\linewidth]{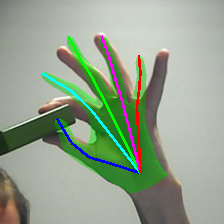}}
\subfloat[(d)]{\includegraphics[width=0.082\linewidth]{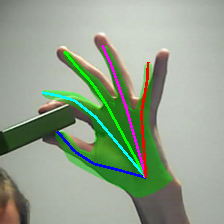}}
\subfloat[(e)]{\includegraphics[width=0.082\linewidth]{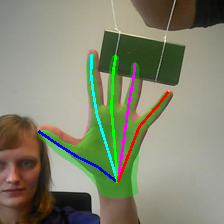}}}
\hspace{-0.055cm}
\subfloat[(\emph{SHD})]{
\subfloat[(a)]{\includegraphics[width=0.082\linewidth]{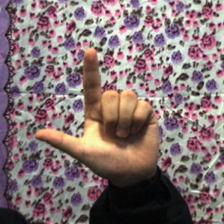}}
\subfloat[(b)]{\includegraphics[width=0.082\linewidth]{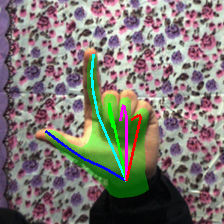}}}
\vspace{-3mm}
\caption{Example dense hand pose estimation results. (\emph{RHD}): (a) input images; (b-c) and (d-e) our results obtained without and with the shape loss $L_\text{\emph{Sh}}$ (Eq.~\ref{loss_sh}), respectively; (b,d) dense hand pose estimation results, and (c,e) estimated shapes in canonical hand pose. (\emph{DO}): (a,c) and (b,d) our results obtained without testing refinement and after applying $20$ iterations of testing refinement, respectively; (e) failure and success cases under occlusion. (\emph{SHD}): (a-b) input images and our results.}
\label{fig:qual_RHD}
\end{figure*}

\paragraph{Experimental settings.} We evaluate the performance of our algorithm on 3 hand pose estimation datasets. Stereo hand pose dataset (\emph{SHD}) provides frames of 12 stereo video sequences each recording a single person performing various gestures ~\cite{stereo_icip2017}. It contains total 36,000 frames. Among them, 30,000 frames sampled from 10 videos constitute a training set while the remaining 6,000 frames (from 2 videos) are used for testing. The rendered hand pose dataset (\emph{RHD}) contains 43,986 synthetically generated images showing 20 different characters performing 39 actions where 41,258 images are provided for training while the remaining 2,728 frames are reserved for testing~\cite{iccv_2017_zimmerman}. Both datasets are recorded under varying backgrounds and lighting conditions and they are provided with the ground-truth 2D and 3D skeleton positions of 21 keypoints (1 for palm and 4 for each finger), on which the accuracy is measured. The Dexter+Object dataset (\emph{DO}) contains 3,145 video frames sampled from 6 video sequences recording a single person interacting with an object (see~Fig.~\ref{fig:qual_RHD}(\emph{DO}))~\cite{hand_eccv_2016_2}. This dataset provides ground-truth 3D skeleton positions for the 5 finger-tips of the left hand: The overall accuracy is measured in these finger-tip locations. 
Following the experimental settings in \cite{iccv_2017_zimmerman,eccv2018_rgbhand,eccv2018_rgbhand2} we train our system on 71,258 frames combining the original training sets of \emph{SHD} and \emph{RHD}. For testing, the remaining frames in \emph{SHD} and \emph{RHD}, respectively and the entire \emph{DO} is used. The overall hand pose estimation accuracy is measured in the area under the curve (AUC) and the ratio of correct keypoints (PCK) with varying thresholds for each~\cite{iccv_2017_zimmerman,eccv2018_rgbhand,eccv2018_rgbhand2}.

\begin{figure*}[t]
\centering
\subfloat[Accuracies on \emph{RHD} ]{\includegraphics[width=0.33\linewidth]{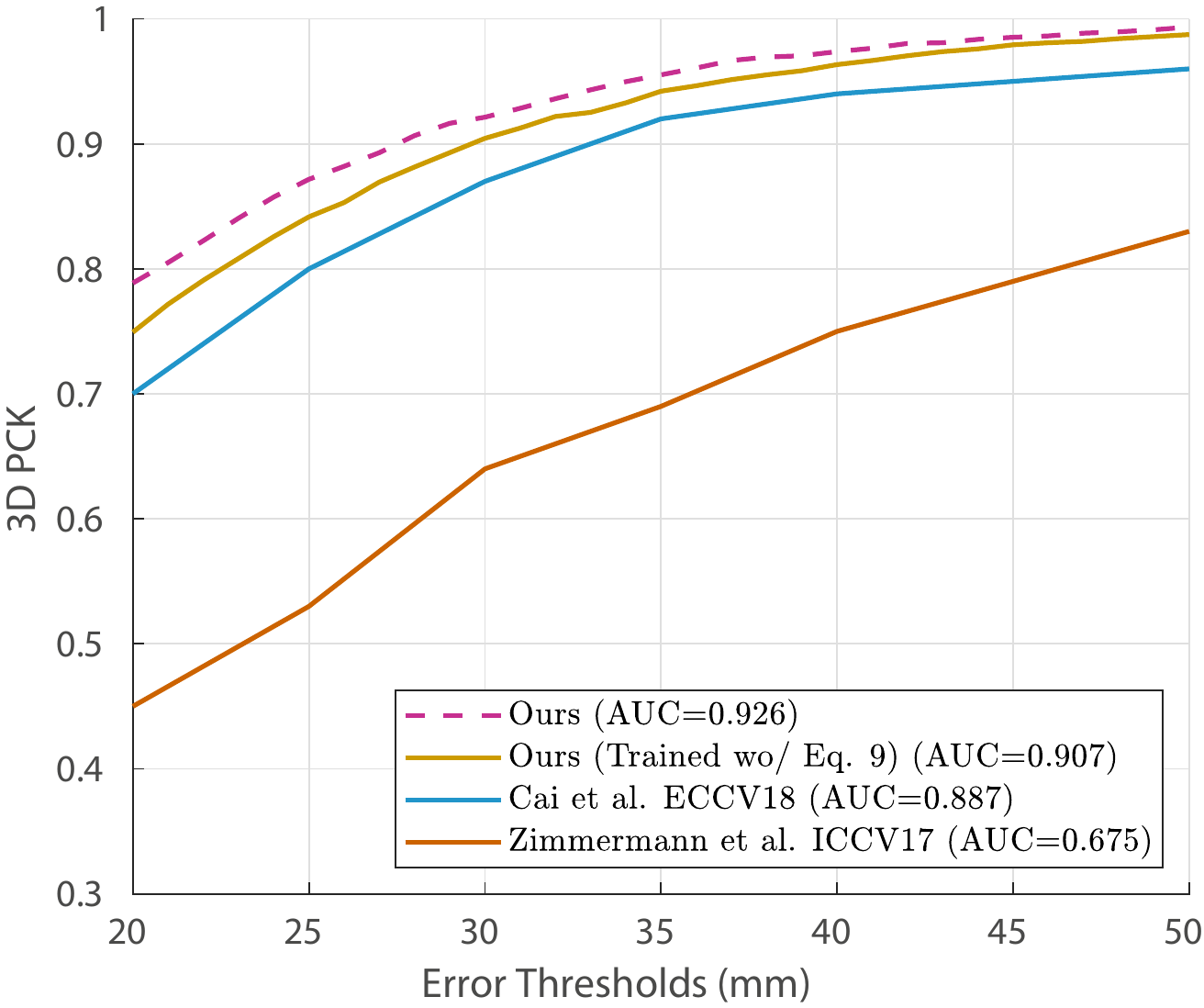}}
\subfloat[Accuracies on \emph{DO} ]{\includegraphics[width=0.33\linewidth]{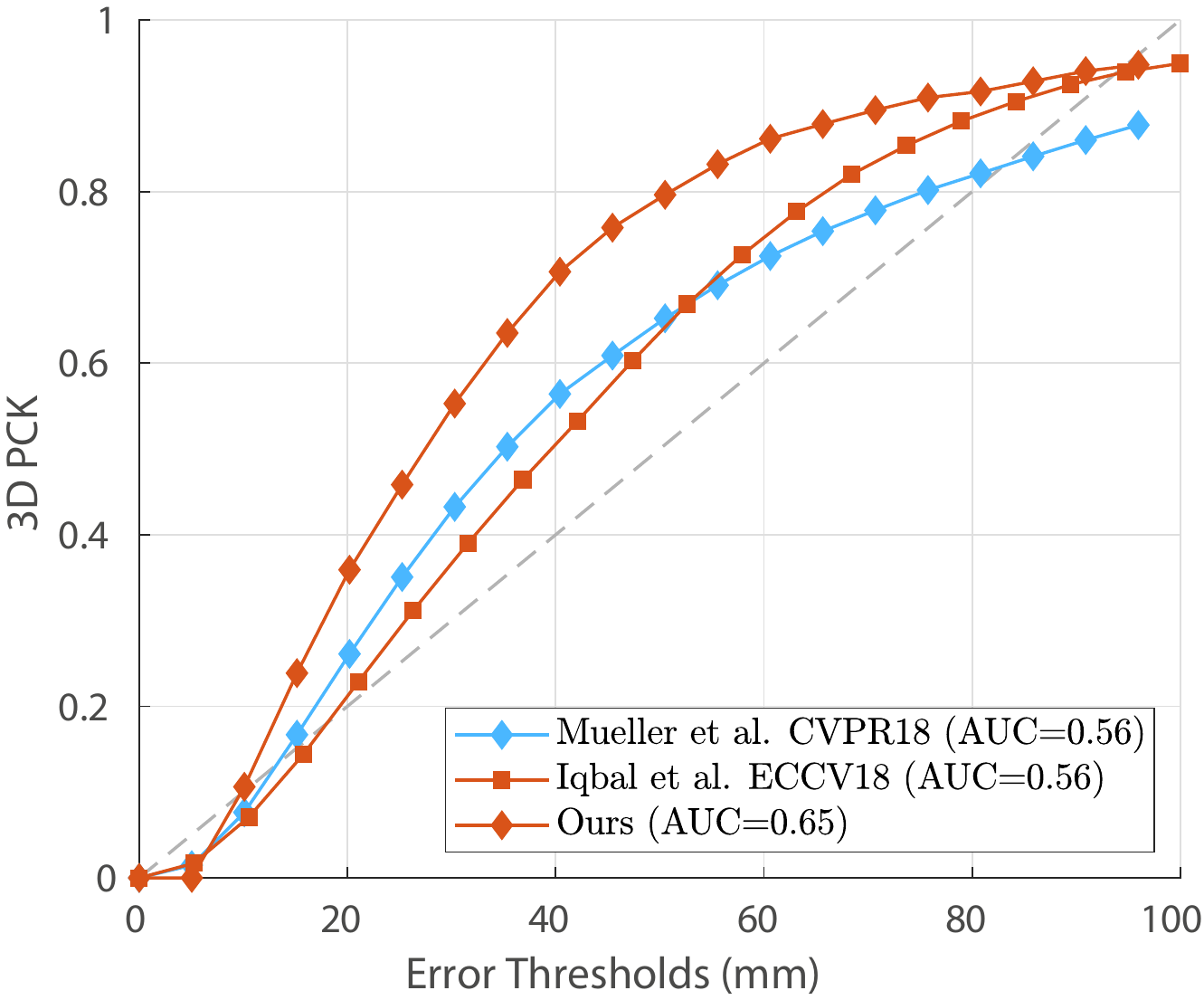}}
\subfloat[Accuracies on \emph{SHD} ]{\includegraphics[width=0.33\linewidth]{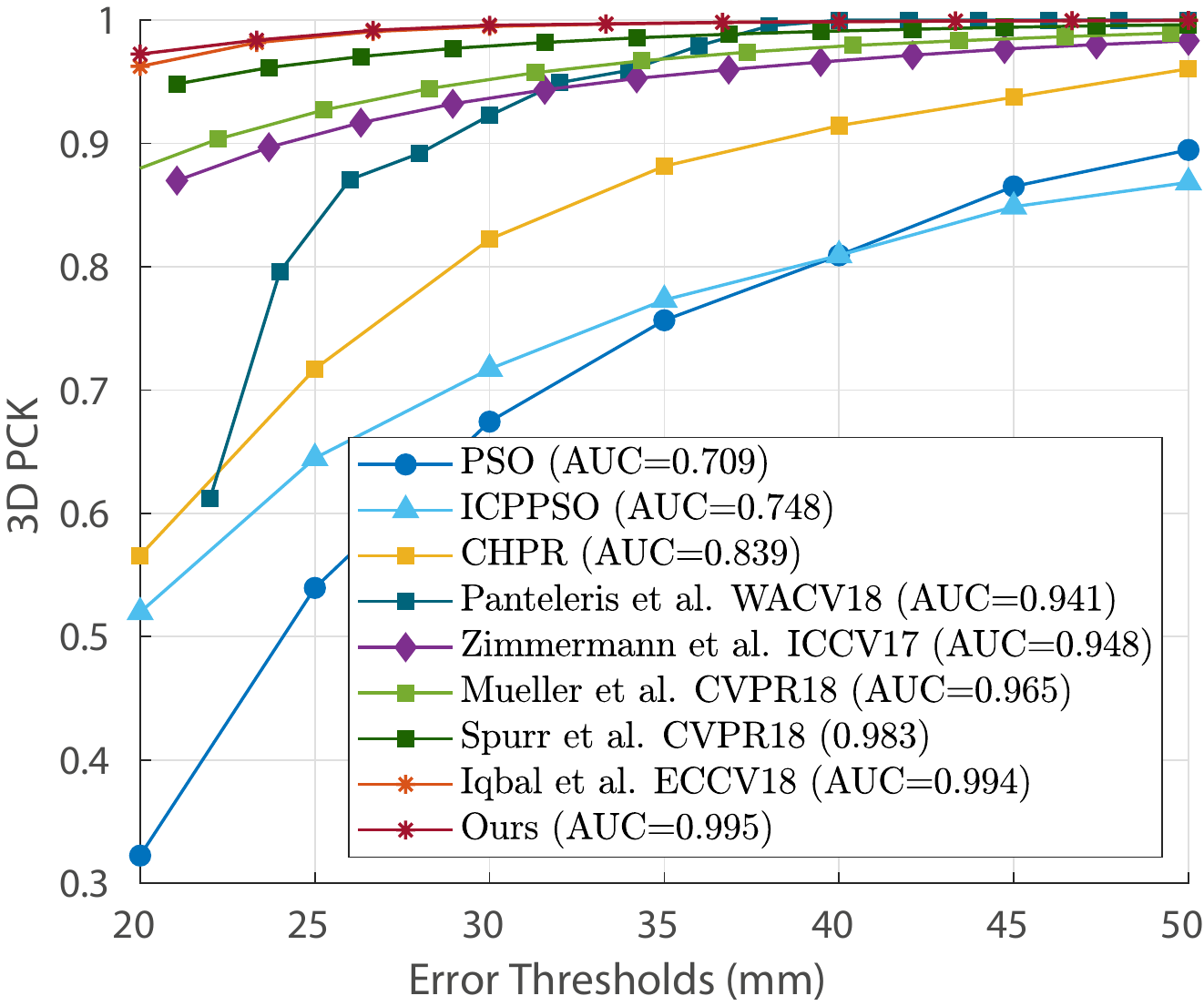}}\\ 
\subfloat[Ablation study results \emph{RHD}]{\includegraphics[width=0.33\linewidth]{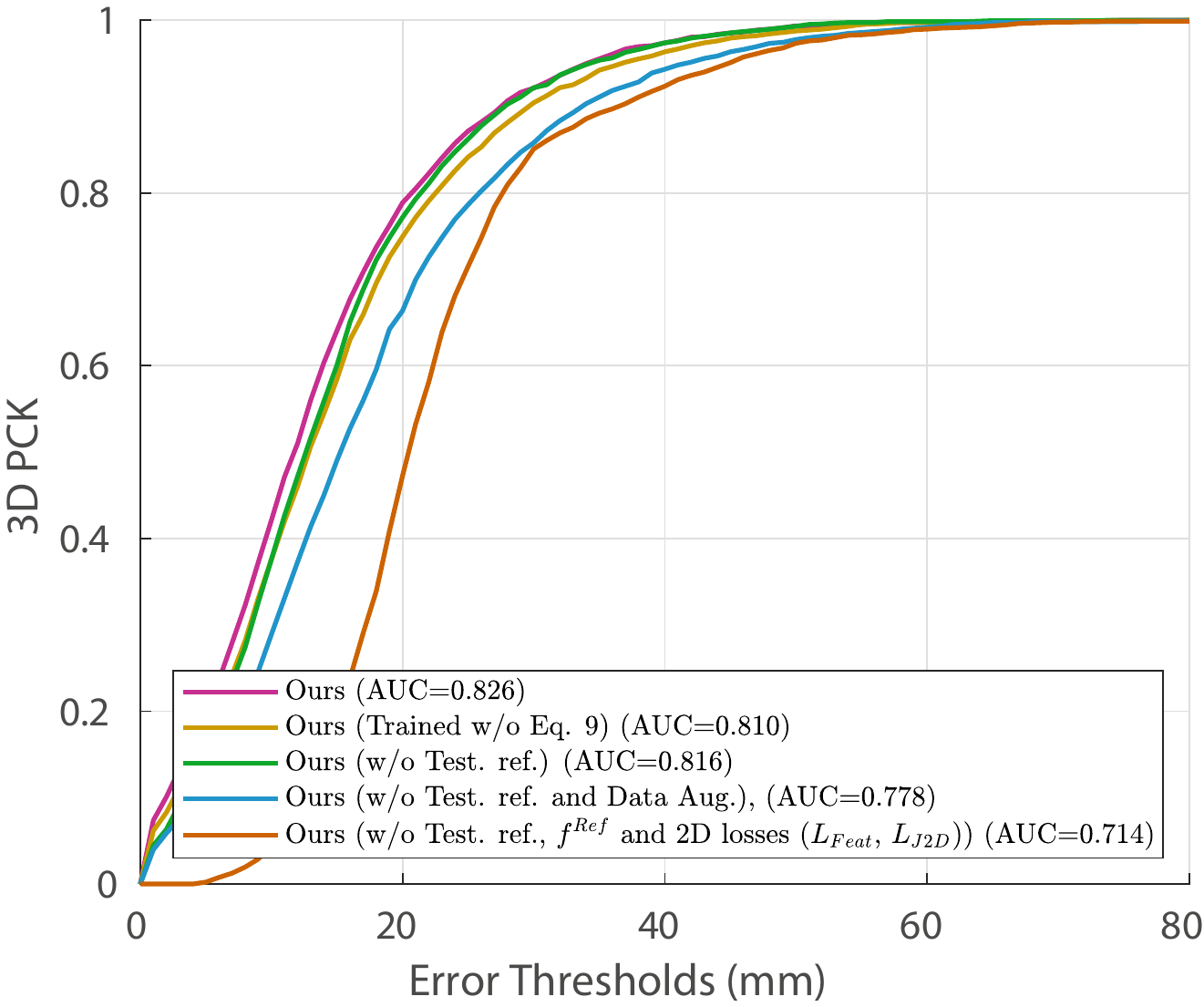}}
\subfloat[Ablation study results \emph{DO}]{\includegraphics[width=0.33\linewidth]{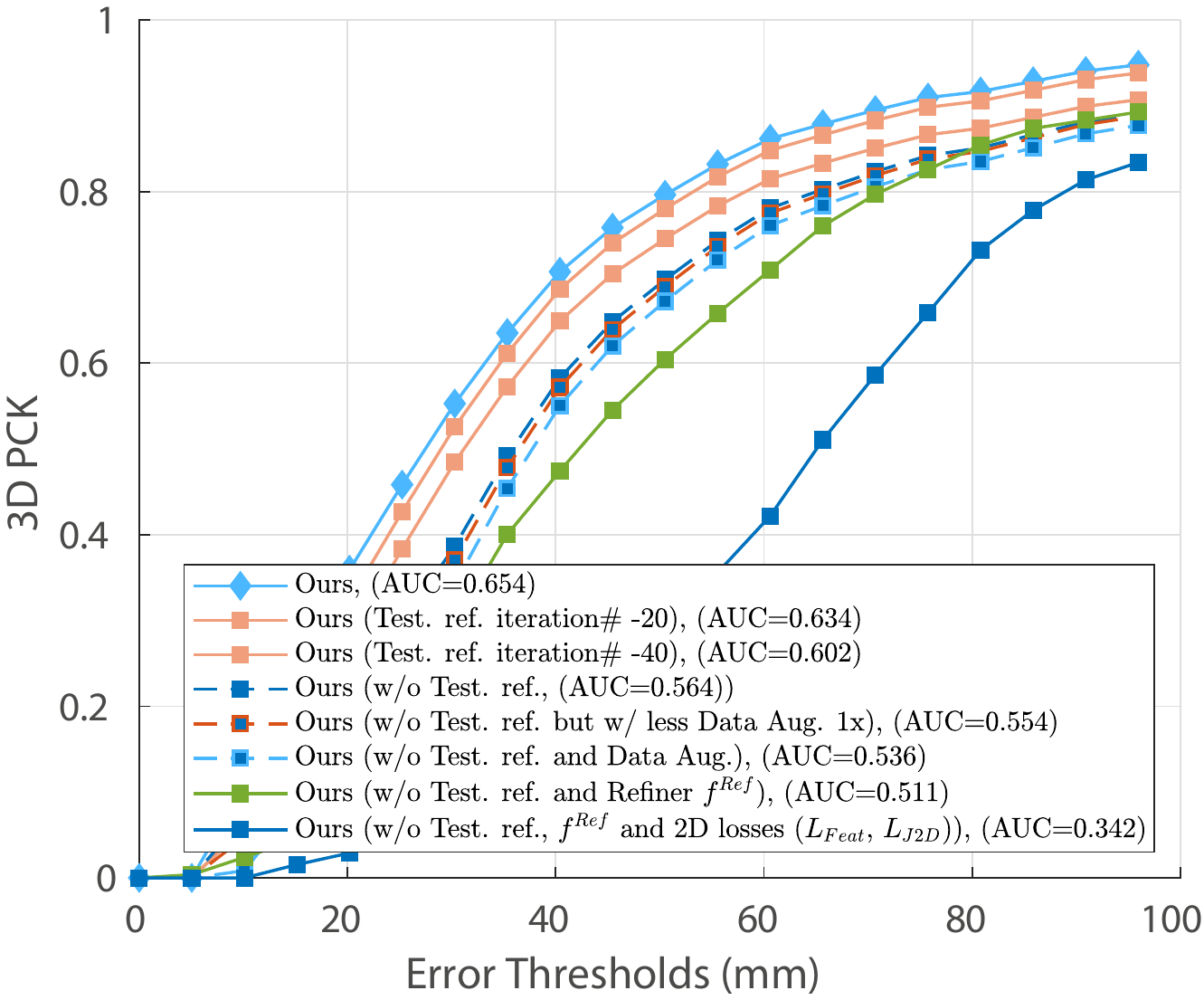}}
\subfloat[Progression of the DHPE accuracies]{\includegraphics[width=0.33\linewidth]{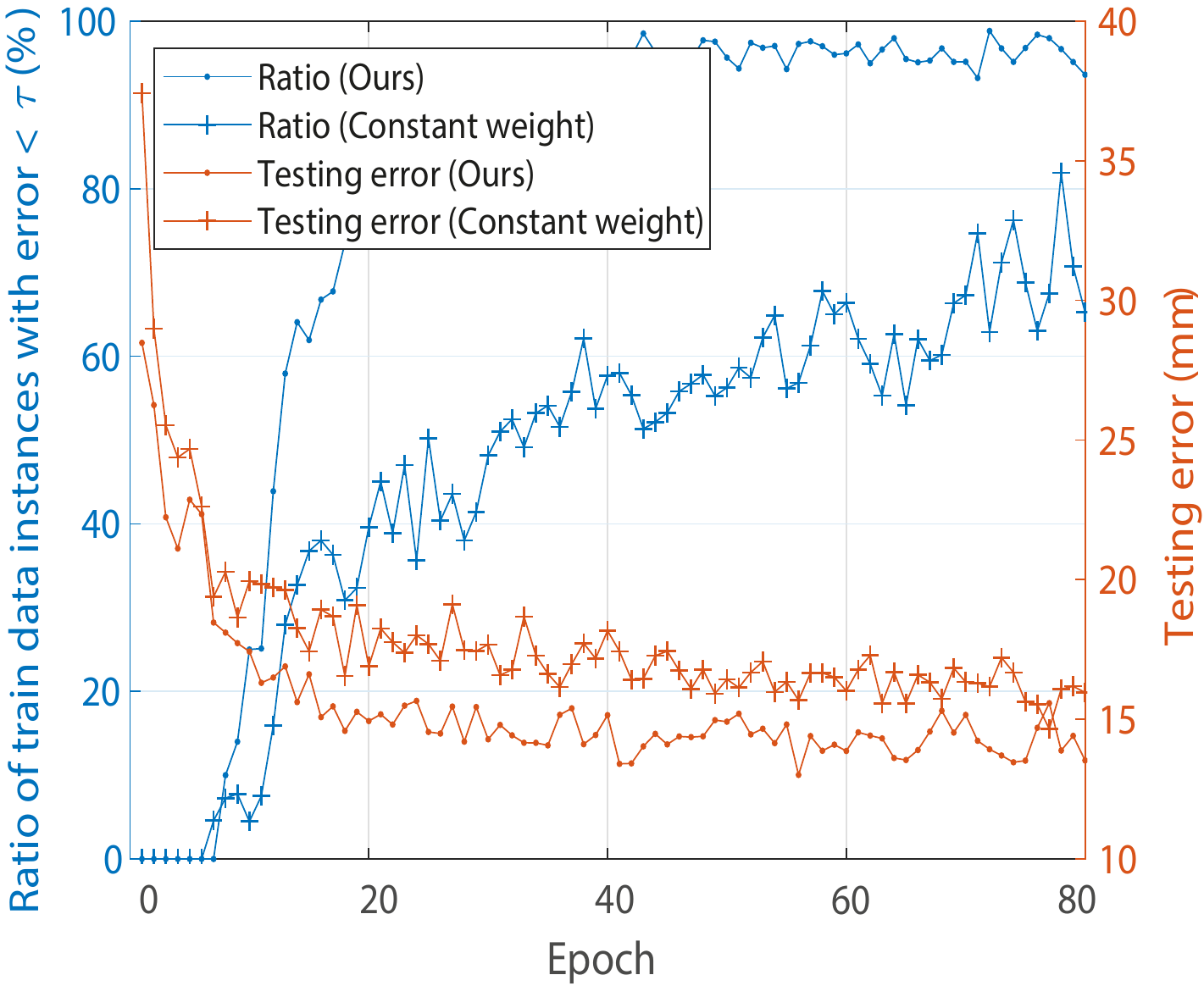}}
\caption{Performances of different algorithms on three benchmark datasets: (a-c) accuracies on \emph{RHD}, \emph{DO}, and \emph{SHD}, respectively; (d-e) evaluation of our algorithm design choices on \emph{RHD} and \emph{DO}, respectively; (f) progressions of the testing errors (orange curves) and the ratio of training data instances with small joint estimation errors ($<\tau$ in Eq.~\ref{eq:hierloss}; blue curves) with $\lambda$ fixed at $0.01$ (curves with dot markers) and with $\lambda$ scheduled based on Eq.~\ref{eq:hierloss} (curves with cross markers).}
\label{fig:results}
\end{figure*}

\begin{table}[t]
\centering
\caption{Performances of different hand segmentation algorithms on \emph{RHD} (higher is better).}
\vspace{-2mm}
\resizebox{1\linewidth}{!}{
\begin{tabular}{|c|cccc|}
\hline
Method & IOU score & Precision & Recall & F1-score\\
\hline
Ours & $65.13\%$ & $82.82\%$ & $75.31\%$ & $78.88$\\
\hline
\cite{iccv_2017_zimmerman} & $35.40\%$ & $36.52\%$ & $92.06\%$ & $52.29$\\
\hline
\cite{cvpr2018_handseg} & $52.68\%$ & $71.65\%$ & $66.55\%$ & $69.00$\\
\hline
\end{tabular}
\label{tab:seg}}
\end{table}

For comparison, we adopt seven hand pose estimation algorithms including five neural networks (CNNs)-based algorithms (\cite{eccv2018_rgbhand,iccv_2017_zimmerman} for \emph{RHD}, \cite{eccv2018_rgbhand2,ganerated_cvpr18} for \emph{DO}, and \cite{ganerated_cvpr18,iccv_2017_zimmerman,cvpr_2018_crossmodal} for \emph{SHD}) and two 3D model fitting-based algorithms \cite{wacv2018_hand,pso}. 

Many existing CNN-based algorithms guide the learning process via building intermediate 2D evidence: 
Zimmermann and Brox's algorithm~\cite{iccv_2017_zimmerman} first estimates 2D skeletons from the input RGB images and thereafter, maps estimated 2D skeletons to 3D skeletons. Spurr~\etal's algorithm~\cite{cvpr_2018_crossmodal} builds a latent space shared by RGB images and 2D/3D skeletons. Cai~\etal's algorithm~\cite{eccv2018_rgbhand} trains an RGB-to-depth synthesizer that generates intermediate depth-map estimates to guide skeleton estimation. 
Similarly, Iqbal~\etal's algorithm~\cite{eccv2018_rgbhand2} reconstructs depth maps from input RGB images. Unlike these algorithms, our algorithm builds full \emph{3D meshes} to guide the skeleton estimation process. 

Mueller~\etal~\cite{ganerated_cvpr18} approached the challenge of building realistic training data pairs of 3D skeletons and RGB images by first generating synthetic RGB images from skeletons and then transferring these images into realistic ones via a GAN transfer network. On the other hand, our algorithm focuses on dense hand pose estimation and thus it generates pairs of RGB images and the corresponding 3D meshes.

\vspace{1mm}
\noindent\textbf{Results.}
Figure~\ref{fig:results} summarizes the results. Our algorithm improved Zimmermann and Brox's algorithm~\cite{iccv_2017_zimmerman} with a large margin (Fig.~\ref{fig:results}(a)). Also, a significant accuracy gain ($\approx$4mm on average) from Cai~\etal's approach~\cite{eccv2018_rgbhand} was obtained, demonstrating the effectiveness of our 3D mesh-guided supervision in comparison to 2.5D depth map-guided supervision. Figure~\ref{fig:results}(a) also demonstrates that jointly estimating the shape and pose (using Eqs.~\ref{loss_art} and \ref{loss_sh}) leads to higher pose estimation accuracy than estimating only the pose. Figure~\ref{fig:results}(b) (\emph{DO}) shows that our algorithm clearly outperforms~\cite{eccv2018_rgbhand2,ganerated_cvpr18}: It should be noted that the comparison with \cite{ganerated_cvpr18} is not fair since their networks were trained on a database tailored for object interaction scenarios as in \emph{DO}. Figure~\ref{fig:results}(c) (\emph{SHD}) further demonstrates that in comparison to the state-of-the-art CNN-based algorithms~\cite{ganerated_cvpr18,iccv_2017_zimmerman,cvpr_2018_crossmodal} as well as 3D model fitting approaches (PSO and \cite{wacv2018_hand}), our algorithm achieves significant performance improvements. The superior performance of our algorithm over \cite{ganerated_cvpr18} shows the effectiveness of our data generation approach. The performance of our algorithm is on par with \cite{eccv2018_rgbhand2} on this dataset but ours outperforms \cite{eccv2018_rgbhand2} on \emph{DO}.

\vspace{1mm}
\noindent\textbf{Ablation study.} 
Figure~\ref{fig:results}(d-e) shows the result of varying design choices in our algorithm on \emph{DO} and \emph{RHD}: The testing refinement step (Sec.~\ref{s:testingrefinement}) had a significant impact on the final performance: The results obtained without this step (`w/o Test. ref.') is considerably worse on \emph{DO}; On \emph{RHD}, results of `w/o Test. ref.' are similar. Our data augmentation strategy (Sec.~\ref{s:dataaugmentation}) further significantly improved the accuracy over `w/o Data Aug.' cases. Figure~\ref{fig:results}(d-e) further show that explicitly providing supervision to our 2D evidence estimator via 2D losses ($L_{J2D}$ and $L_{Feat}$; Eqs.~\ref{loss_feat}-\ref{loss_j2d}) constantly improve the overall accuracy over `w/o 2D loss' cases which correspond to Kanazawa~\etal's iterative estimation framework~\cite{e2eshapepose}. Figure~\ref{fig:results}(f) visualizes the effectiveness of our hierarchical shape and articulation optimization strategy (Eq.~\ref{eq:hierloss}): In comparison to the case of constant $\lambda$ value (`Constant weight'), automatically scheduling $\lambda$ based on Eq.~\ref{eq:hierloss} led to larger fraction of training instances that have small articulation errors ($<\tau$; Eq.~\ref{eq:hierloss}) which then led to faster decrease of test error. 
\vspace{1mm}

\noindent \textbf{Qualitative evaluation.} For qualitative evaluation, we show example dense estimation results in Fig.~\ref{fig:qual_RHD}. Figure~\ref{fig:qual_RHD}(\emph{RHD}) demonstrates the importance of the shape loss $L_\text{\emph{Sh}}$ (Eq.~\ref{loss_sh}) conforming the quantitative results of Fig.~\ref{fig:results}(a,d).
\vspace{1mm}

\noindent \textbf{Hand segmentation performance.}
Systematically evaluating the performance of dense hand pose estimation (including shape) is challenging due to the lack of ground-truth labels. Therefore, we assess the performance indirectly based on the 2D hand segmentation accuracy. Table~\ref{tab:seg} shows the results measured in 4 different object segmentation performance criteria. In comparison to the state-of-the-art hand segmentation networks~\cite{cvpr2018_handseg,iccv_2017_zimmerman} (note \cite{cvpr2018_handseg} is fine-tuned for \emph{RHD}), our algorithm led to significantly higher accuracy. Also, Fig.~\ref{fig:segresult} shows that the state-of-the-art segmentation net~\cite{cvpr2018_handseg} is distracted by other skin colored objects than hands (\eg arms) while our algorithm, by explicitly estimating 3D meshes, can successfully disregarded these distracting backgrounds.

\section{Conclusions}
We have presented dense hand pose estimation (DHPE) network, a CNN-based framework that reconstructs 3D hand shapes and poses from single RGB images. DHPE decomposes into the 2D evidence estimator, 3D mesh estimator and projector. The projector, via the neural renderer, replaces insufficient full 3D supervision with indirect supervision by 2D segmentation masks/3D joints, and enables generating new data. In the experiments, we have demonstrated that stratifying 2D/3D estimators improves accuracy, 
updating 2D skeletons helps self-supervision in the iterative testing refinement and improves 3D skeleton estimation, and jointly estimating 3D hand shapes and poses offers the state-of-the-art accuracy in both 3D hand skeleton estimation and 2D hand segmentation tasks.

{\small
\bibliographystyle{ieee_fullname}
\bibliography{egbib}
}

\end{document}